\documentclass[letterpaper, 10 pt, conference]{ieeeconf}  % Comment this line out

\IEEEoverridecommandlockouts                              % This command is only
\usepackage[colorlinks,linkcolor=green,citecolor=red,urlcolor=blue,bookmarks=true,hypertexnames=true]{hyperref} 
\usepackage[utf8]{inputenc}
\usepackage[T1]{fontenc}
\usepackage{balance}

\usepackage{epsfig} % for postscript graphics files
\usepackage{mathptmx} % assumes new font selection scheme installed
\usepackage{mathptmx} % assumes new font selection scheme installed
\usepackage{amsmath} % assumes amsmath package installed
\usepackage{amssymb}  % assumes amsmath package installed
\usepackage{booktabs}
\usepackage{url}
\usepackage{multirow}
\usepackage{multicol}
\usepackage{color}

\usepackage{tabu}                     % 表格插入
\usepackage{multirow}                 % 一般用以设计表格，将所行合并
\usepackage{multicol}                 % 合并多列
\usepackage{multirow}                % 合并多行
\usepackage{float}                    % 图片浮动
\usepackage{makecell}                 % 三线表-竖线
\usepackage{booktabs} 
\usepackage{threeparttable}

\usepackage{bbm}

\title{\LARGE \bf A Concise Survey on Lane Topology Reasoning for HD Mapping}

\author{Yi Yao$^{1,\dagger}$, Miao Fan$^{2,\dagger,*}$, Shengtong Xu$^{3}$,  Haoyi Xiong$^{4}$, Xiangzeng Liu$^{5}$, Wenbo Hu$^{6}$, and Wenbing Huang$^{7}$    % <-this % stops a space
\thanks{$^{1}$Senior engineer at NavInfo Co., Ltd., China.}
\thanks{$^{2}$Chief scientist at NavInfo Co., Ltd., China. Senior member of IEEE.}
\thanks{$^{3}$Principal product manager at Autohome Inc., China.}
\thanks{$^{4}$Principal scientist at Baidu Inc., China. Senior member of IEEE.}
\thanks{$^{5}$Associate professor at Xidian University of Technology, China.}
\thanks{$^{6}$Associate professor at Hefei University of Technology, China.}
\thanks{$^{7}$Associate professor at Renmin University of China.}
\thanks{$\dagger$Equal contribution.}
\thanks{*Correspondence: {miao.fan@ieee.org}.}
}

\begin{document}

\maketitle
\thispagestyle{empty}
\pagestyle{empty}

%%%%%%%%%%%%%%%%%%%%%%%%%%%%%%%%%%%%%%%%%%%%%%%%%%%%%%%%%%%%%%%%%%%%%%%%%%%%%%%%
\begin{abstract}
Lane topology reasoning techniques play a crucial role in high-definition (HD) mapping and autonomous driving applications. While recent years have witnessed significant advances in this field, there has been limited effort to consolidate these works into a comprehensive overview. This survey systematically reviews the evolution and current state of lane topology reasoning methods, categorizing them into three major paradigms: procedural modeling-based methods, aerial imagery-based methods, and onboard sensors-based methods. We analyze the progression from early rule-based approaches to modern learning-based solutions utilizing transformers, graph neural networks (GNNs), and other deep learning architectures. The paper examines standardized evaluation metrics, including road-level measures (APLS and TLTS score), and lane-level metrics (DET and TOP score), along with performance comparisons on benchmark datasets such as OpenLane-V2. We identify key technical challenges, including dataset availability and model efficiency, and outline promising directions for future research. This comprehensive review provides researchers and practitioners with insights into the theoretical frameworks, practical implementations, and emerging trends in lane topology reasoning for HD mapping applications.
\end{abstract}
{\keywords Topology reasoning, high-definition maps, road network, deep learning, procedural modeling, aerial imagery, onboard sensors.}

%%%%%%%%%%%%%%%%%%%%%%%%%%%%%%%%%%%%%%%%%%%%%%%%%%%%%%%%%%%%%%%%%%%%%%%%%%%%%%%%

\section{Introduction}
\begin{figure}
\centering
  \includegraphics[width=0.95\columnwidth]{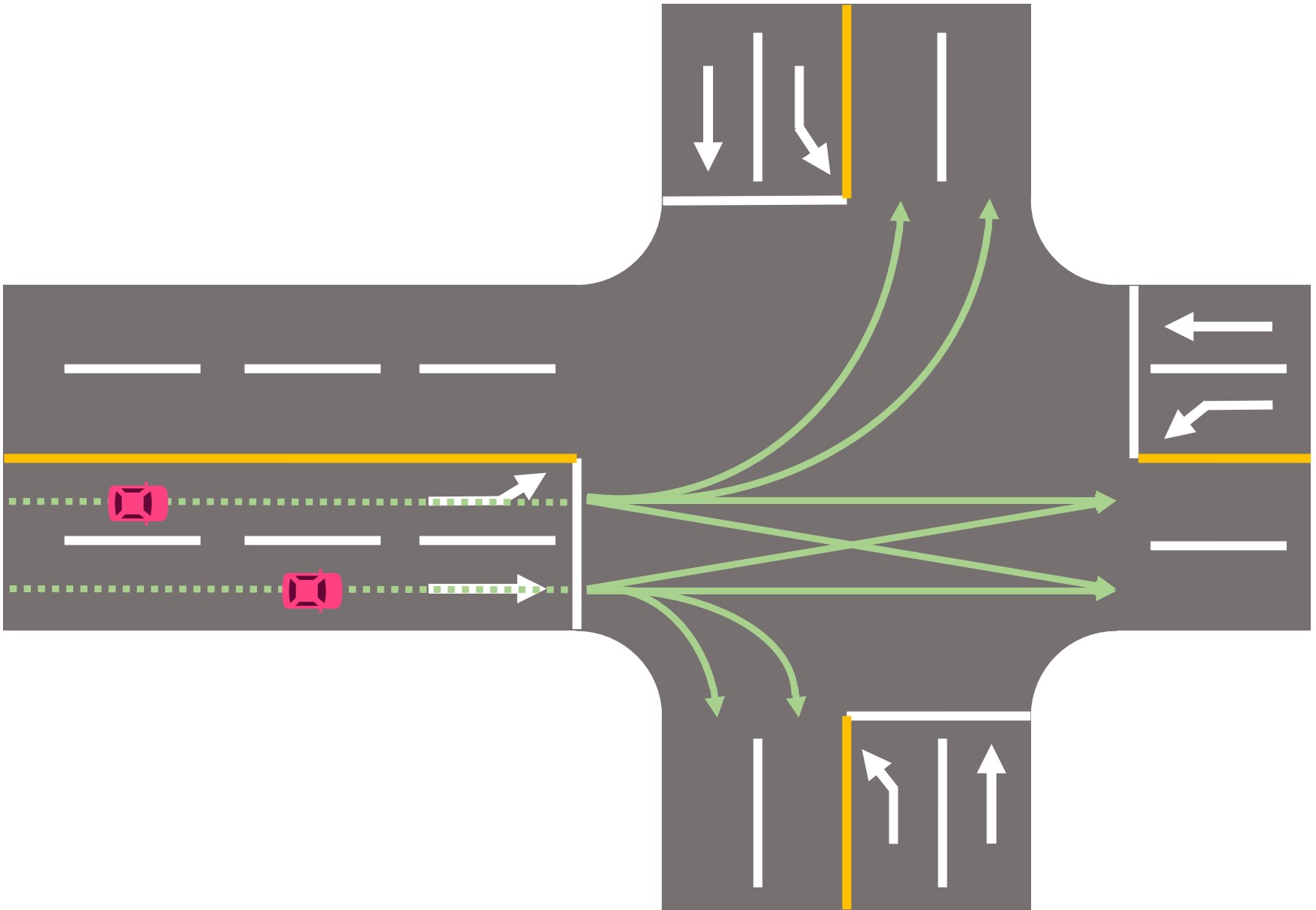}
  \caption{Schematic diagram of lane topology at an intersection.}
  \label{fig:topo}
\end{figure}

High-definition (HD) maps are fundamental for autonomous driving, providing detailed representations of road networks, lane configurations, and traffic elements \cite{bao2023review,chen2022milestones,r0}. Among the various aspects of HD mapping, lane topology reasoning - the understanding of how different lanes connect and relate to each other (see Fig. \ref{fig:topo}) - is particularly crucial for safe and efficient autonomous navigation \cite{elghazaly2023high}. This field has evolved significantly over the past two decades, progressing from simple rule-based approaches to sophisticated deep-learning methods.

\begin{figure}
\centering
  \includegraphics[width=0.95\columnwidth]{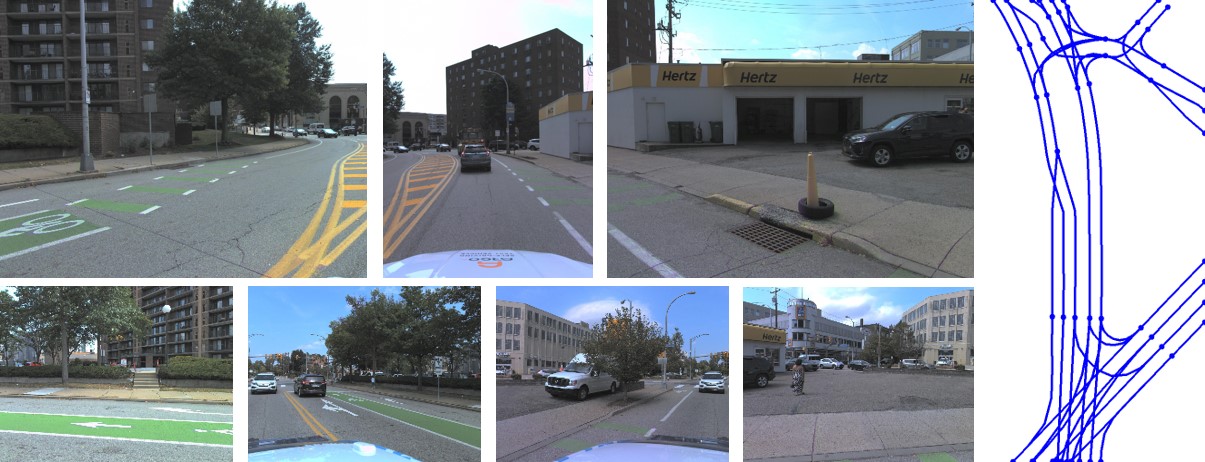}
  \caption{Illustration of lane topology reasoning for high-definition (HD) mapping, adapted from OpenLane-V2 \cite{wang2024openlane}. This example demonstrates the transformation from multi-perspective onboard camera images to a comprehensive bird's-eye view (BEV) lane topology representation.}
  \label{fig:1}

\end{figure}

\begin{figure*}[htbp]
\centering
\includegraphics[width=\textwidth]{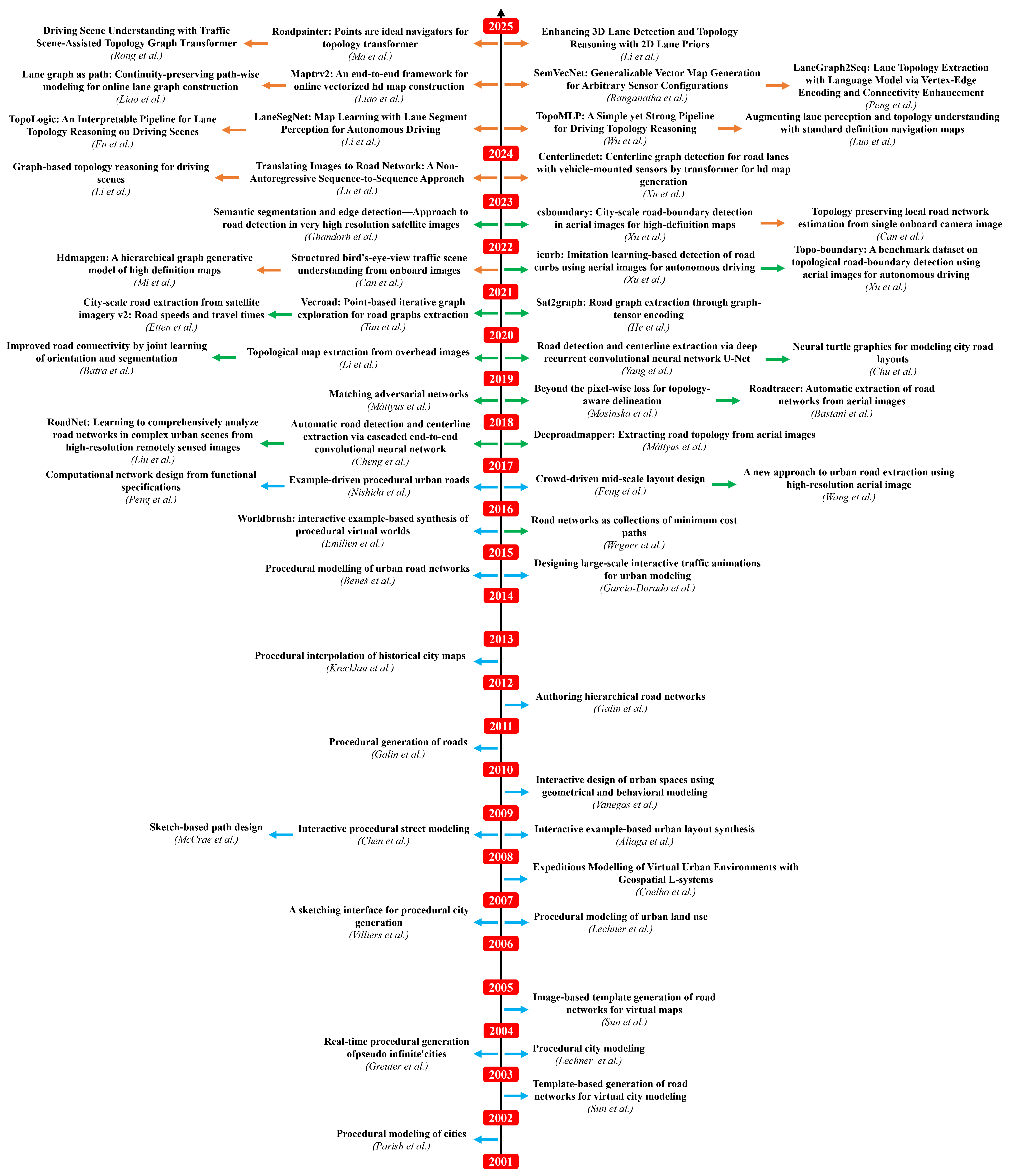}
\caption{The evolution of lane topology reasoning research from 2001 to 2024, illustrating the field's methodological progression and technological transitions. The timeline tracks three distinct approaches: procedural modeling-based methods (blue arrows), aerial imagery-based methods (green arrows), and onboard sensors-based methods (orange arrows). The distribution of arrows reveals the field's transformation from early procedural modeling techniques to modern vision-based approaches, with a notable acceleration in research activity post-2020. This visualization highlights the parallel development of different methodologies and the increasing focus on onboard sensors-based solutions in recent years.}
\label{fig:time}
\end{figure*}

The development of lane topology reasoning methods can be broadly categorized into three major paradigms. Early approaches relied on procedural modeling techniques, utilizing computer graphics and rule-based systems to generate road networks. While foundational, these methods were limited by their reliance on manually designed rules and lack of flexibility. The advent of aerial imagery-based methods marked a significant shift, introducing computer vision techniques to extract road topology from overhead views. However, these approaches often struggled with occlusions caused by buildings and trees, and could only extract coarse, road-level networks. In recent years, we have witnessed the emergence of onboard sensors-based methods (see Fig. \ref{fig:1}), which represent the current state-of-the-art in lane topology reasoning field. These approaches leverage deep learning architectures, typically consisting of several key components: a backbone network for feature extraction (such as ResNet \cite{he2016deep} or Swin Transformer \cite{liu2021swin}), perspective transformation for bird's-eye view conversion, lane centerline detection, and topology reasoning modules. Modern methods often incorporate multiple input modalities, including cameras, LiDAR, and existing SD maps~\cite{r00}, to enhance performance. The introduction of transformer-based architectures and graph neural networks (GNNs) has further advanced the field, enabling more sophisticated topology reasoning capabilities.

% Despite these advances, several challenges remain. The availability of comprehensive datasets for training and evaluation, the computational efficiency of deep learning models, and the robustness of topology reasoning in complex scenarios continue to be active areas of research. Additionally, the integration of various data sources and the development of more efficient architectures remain important objectives for practical applications.

Existing surveys \cite{bao2023review,chen2022milestones,elghazaly2023high} have detailed HD map creation and lane topology construction, but they lack a systematic presentation of the latest research. This survey provides a comprehensive review of lane topology reasoning methods, analyzing their evolution, capabilities, and limitations. We examine different evaluation metrics used in the field, including road-level and lane-level metrics, and compare the performance of various approaches on standard benchmarks. Furthermore, we discuss current challenges and identify promising directions for future research, aiming to provide researchers and practitioners with valuable insights into this rapidly evolving field.

% The remainder of this paper is organized as follows: Section II reviews the three major paradigms in detail. Section III discusses commonly used datasets and also examines evaluation metrics and performance comparisons.  section IV concludes with challenges and future directions.

% The remainder of this paper is organized as follows: Section II reviews the three major paradigms in detail. Section III discusses commonly used datasets and also examines evaluation metrics and performance comparisons. Section IV points out the challenges and future directions. Finally, we conclude in section V.

The remainder of this paper is structured as follows: Section II details the three major paradigms; Section III discusses common datasets, evaluation metrics, and performance comparisons; Section IV identifies challenges and future directions; and Section V concludes.

\section{Review on Methodology}
% In this section, we provide a comprehensive review of representative methods for lane topology reasoning proposed up to 2024. As shown in Fig. \ref{fig:time}, based on the developmental paradigms, these methods are categorized into three primary groups. Before 2017, procedural modeling-based techniques dominated this field. Aerial imagery-based approaches received more attention from 2017 to 2022. Since 2023, the research trend has shifted towards solutions based on onboard sensors.

This section reviews representative lane topology reasoning methods published through 2024. As illustrated in Fig. \ref{fig:time}, these methods are categorized into three groups based on their developmental paradigms. Procedural modeling dominated before 2017, and aerial imagery-based approaches gained prominence from 2017 to 2022. Since 2023, onboard sensors-based solutions have become increasingly prevalent.

\subsection{Procedural Modeling-Based Methods}

\begin{figure*}
\centering
  \includegraphics[width=0.95\textwidth]{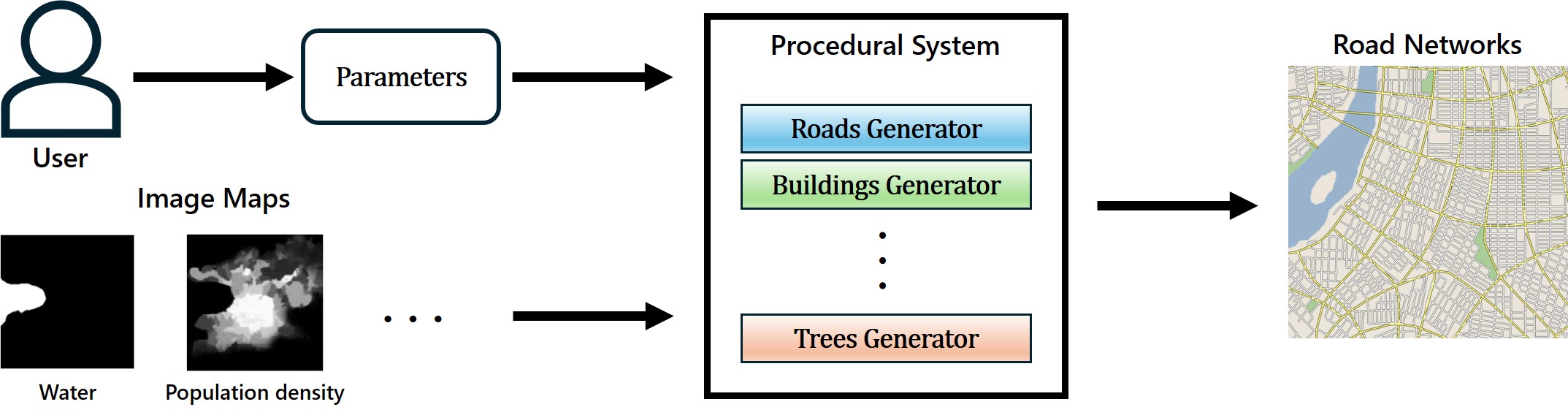}
  \caption{The pipeline of procedural modeling-based methods.}
  \label{fig:pipline1}
\end{figure*}

Early explorations in lane topology reasoning predominantly employed computer modeling techniques to simulate urban road networks, and the high ability to amplify data is the main merit. The workflow of these methods is depicted in Fig. \ref{fig:pipline1}. After inputting 2D image maps of the city (e.g., geographical maps and statistical maps), these approaches generated virtual urban maps that incorporated buildings, road networks, and other urban features. Interactive parameter tuning enabled users to refine the outcomes according to specific requirements.

Parish et al. \cite{parish2001procedural} presented pioneering work by applying the L-system to generate urban road networks, this concept was further extended in subsequent studies \cite{coelho2007expeditious}. \cite{sun2002template,greuter2003real,sun2004image} summarized some basic templates to generate road networks by overlaying different templates. Meanwhile, Chen et al. \cite{chen2008interactive} introduced tensor field representations to model road geometries. Later advancements incorporated additional constraints and functionalities. Terrain-aware approaches \cite{galin2010procedural, benevs2014procedural, galin2011authoring} ensured geographical consistency, while intelligent simulation techniques \cite{krecklau2012procedural, lechner2003procedural, lechner2006procedural, feng2016crowd, peng2016computational, vanegas2009interactive} enhanced the realism of traffic modeling. Furthermore, sketch-based methods \cite{de2006sketching, mccrae2008sketch} enabled intuitive user interactions, and example-driven approaches \cite{aliaga2008interactive, emilien2015worldbrush, nishida2016example} leveraged pre-existing data to expedite road network generation. Traffic-related factors, such as road occupancy, travel time, and emission impacts, were explicitly addressed in \cite{garcia2014designing}.

\subsection{Aerial Imagery-Based Methods}

\begin{figure*}
\centering
  \includegraphics[width=0.95\textwidth]{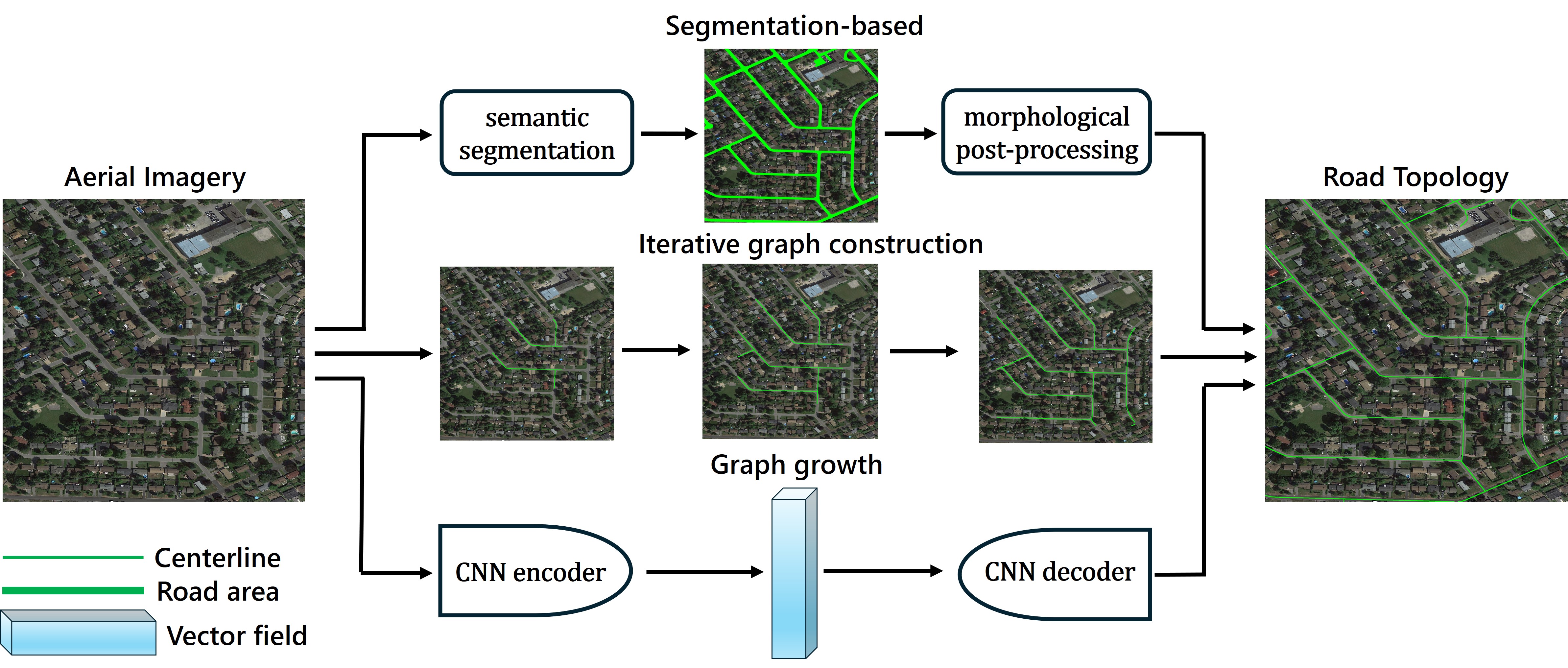}
  \caption{The pipeline of aerial imagery-based methods.}
  \label{fig:pipline2}
\end{figure*}

While procedural modeling-based methods were foundational in this field, these methods still had limited controllability and relied heavily on artificially designed rules. In addition, evaluating the influence of these rules and parameters on the final output is inherently challenging, requiring extensive iterative experimentation to achieve satisfactory results. Consequently, these methods lack flexibility. In contrast, the advent of deep learning has revolutionized HD mapping by enabling the direct extraction of road topology from aerial imagery. Leveraging neural networks, these approaches minimize manual intervention and streamline the mapping process. The general workflow of these methods is illustrated in Fig. \ref{fig:pipline2}.

Several methods \cite{mattyus2017deeproadmapper,cheng2017automatic,mosinska2018beyond,mattyus2018matching,batra2019improved,ghandorh2022semantic,wegner2015road,etten2020city,yang2019road,wang2016new,liu2018roadnet} leverage semantic segmentation networks to identify road areas in aerial imagery, followed by morphological post-processing to extract road topology. Alternative approaches involve iterative graph construction \cite{bastani2018roadtracer,xu2021topo,chu2019neural,li2019topological,tan2020vecroad,xu2021icurb} and graph growth \cite{he2020sat2graph,xu2022csboundary} methods. Iterative graph construction methods are initiated by identifying several vertices from the road network and progressively expanding the graph by iteratively adding new vertices and edges based on predefined decision functions. In contrast, graph generation methods are more direct. These approaches employ the vector field encoding strategy, wherein aerial imagery is first encoded into latent representations. Neural networks are subsequently used to predict graph vertices, followed by decoding processes predicting the adjacency matrix and reconstructing the road network.

\subsection{Onboard Sensors-Based Methods}

% \begin{figure*}[htbp]
%     \centering
%     \includegraphics[width=0.95\textwidth]{figures/pipeline3.jpg}
%     \caption{The pipeline of onboard camera-based methods.}
%     \label{fig:pipline3}
% \end{figure*}

\begin{figure*}[htbp]
    \centering
    \includegraphics[width=0.95\textwidth]{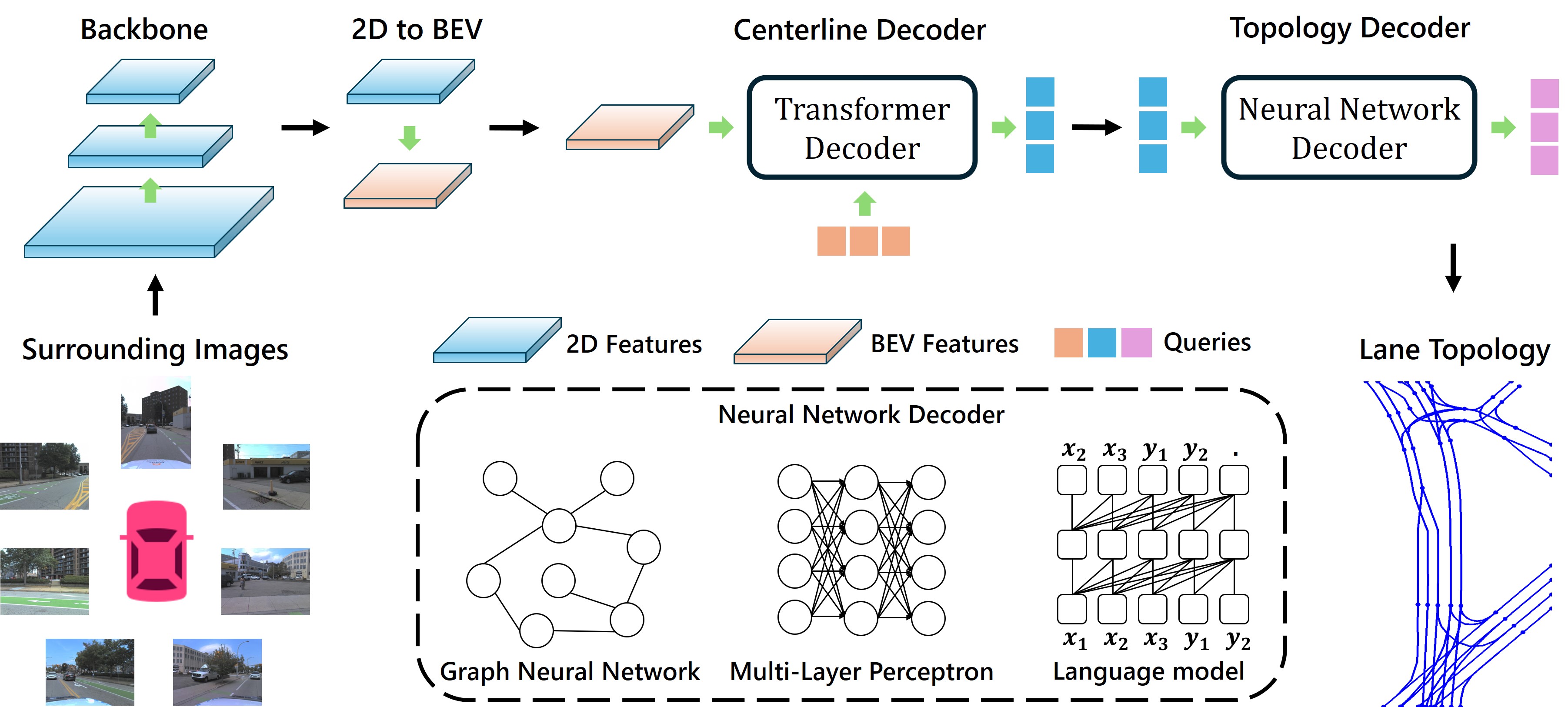}
    \caption{The pipeline of onboard sensors-based methods. Although some methods used multimodal input data, almost all contained images from onboard cameras. Therefore, for simplicity, the input data presented in the figure contains only onboard camera images.}
    \label{fig:pipline3}
\end{figure*}

Even though aerial imagery-based methods have introduced novel approaches to this field, they are limited to extracting coarse, road-level networks. Furthermore, occlusions caused by buildings and trees often render road areas incomplete in aerial imagery, affecting the accuracy of results. Therefore, these methods are unsuitable for the automatic construction of HD maps or action planning of autonomous vehicles. To address these limitations, onboard sensors-based methods have emerged, utilizing unimodal or multimodal data captured from vehicle-mounted sensors to construct detailed lane topology around the ego-vehicle. These methods represent a significant shift in HD mapping, offering enhanced accuracy and applicability. A typical pipeline for onboard sensors-based methods is illustrated in Fig. \ref{fig:pipline3}, including the backbone, perspective transformation, lane decoder, and topology decoder. The backbone is usually composed of feature extraction networks such as ResNet \cite{he2016deep} and Swin Transformer \cite{liu2021swin}. Perspective transformation then converts these features into a bird's-eye view (BEV) format. After obtaining the BEV features, a lane centerline decoder is usually designed using the transformer-based architecture to output the lane centerline features based on the information requirements. In this part, a set of learned query vectors is generally used for mapping different centerline instances. The function of the topology decoder is to establish a connection between these lane centerline instances, normally using GNNs, MLPs, and language models. In addition to onboard camera images, some approaches augment input features with supplementary data sources, including SD maps \cite{luo2024augmenting,ma2024roadpainter,fu2024topologic}, LiDAR point clouds \cite{xu2023centerlinedet,10588555,liao2024maptrv2}, and traffic elements (e.g., traffic lights and traffic boards) \cite{li2023graph,rong2024driving,wu2023topomlp,li2024enhancing}, thereby improving performance.

Can et al. \cite{can2021structured} proposed one of the first works, using a single image from a front-facing camera as input and employing a transformer-based architecture, which is similar to DETR \cite{carion2020end}, to predict centerline instances. This method employs additional MLPs to output the connectivity of centerline instances. Likewise, these methods \cite{li2024lanesegnet,wu2023topomlp,li2024enhancing,ma2024roadpainter,jia2024lanedag} also use a set of MLPs to predict the topology relationship. Expanding upon \cite{can2021structured}, Can et al. \cite{can2022topology} introduced the concept of the minimal ring and coverage accurately estimates lane connectivity in intersection areas. LaneGAP \cite{liao2024lane} further refined topology estimation by employing the Path2Graph algorithm to recover the topology relationships. HDMapGen \cite{mi2021hdmapgen} autoregressively generates lane topology using only graph-based models. Some solutions \cite{xu2023centerlinedet,fu2024topologic,li2023graph,rong2024driving,liao2024maptrv2} first detect lane centerlines and map each lane centerline to a vertex, then update the lane topology using a graph-based model. Additionally, methods such as \cite{lu2023translating,peng2024lanegraph2seq} serialize lane centerline features and use language models to establish topology relationships.

\subsection{Other Methods}

Although this survey focuses on the three method types described above, other notable approaches exist. For example, Jia et al. \cite{jia2024lanedag} generated lane topology from lane lines and road boundaries in existing HD maps, and He et al. \cite{he2018roadrunner} and Stanojevic et al. \cite{stanojevic2018robust} extracted lane topology from GPS trajectories. While valuable, these methods fall outside the scope of this review.

% Besides the three types of relevant methods described above, there are other excellent efforts. For instance, \cite{jia2024lanedag} generated lane topology based on lane lines and road boundaries from existing HD maps, \cite{he2018roadrunner,stanojevic2018robust} extracted lane topology from GPS trajectories. Although these approaches are worth exploring, they are not the focus of this survey.

\section{Datasets and Benchmarks}

% In this section, we discuss the associated datasets and evaluation metrics

In this section, we list datasets that are commonly employed in the methods discussed above. Additionally, we examine different evaluation metrics used in the field and compare the performance of various approaches on benchmark datasets. Due to space constraints, the selection is not exhaustive.

\subsection{Datasets}
Relevant datasets can be categorized into 2D datasets and 3D datasets. These datasets, summarized in Table \ref{tab:1}, play a critical role in advancing road topology extraction and lane topology reasoning by offering diverse and high-quality data tailored to specific research needs.

\subsubsection{2D Datasets}

To extract road topology from aerial images, several 2D datasets have been utilized, including SpaceNet \cite{van2018spacenet}, DeepGlobe \cite{demir2018deepglobe} and Topo-boundary \cite{xu2021topo}. SpaceNet is a large-scale dataset containing annotations for building footprints and road networks, widely used in urban topology extraction tasks. DeepGlobe focused on rural areas with diverse land cover types, this dataset offers high-density road network annotations, making it suitable for tasks requiring detailed mapping in non-urban environments. Topo-boundary is specifically designed for offline topological road-boundary detection, enabling the development of methods targeting precise boundary delineation. In addition, procedural modeling-based methods often utilize 2D image maps generated by plotting or scanning statistical and geographical maps of cities. 

\subsubsection{3D Datasets}

nuScenes \cite{caesar2020nuscenes}, Argoverse2 \cite{wilson2argoverse} and OpenLane-V2 \cite{wang2024openlane} are significant datasets to be utilized in onboard sensors-based methods. Both nuScenes and Argoverse2 are composed of data from high-resolution onboard sensors, including cameras and LiDAR. OpenLane-V2 is the unique 3D dataset specifically designed for lane topology reasoning tasks, and it is divided into two subsets: subset\_A and subset\_B, which are derived from Argoverse2 and nuScenes, respectively. These subsets ensure comprehensive coverage of scenarios relevant to lane topology reasoning.

\begin{table*}[h]
\centering
\caption{Comparison of current relevant datasets.}
\begin{tabular}{l|rrrrrrr}
\toprule
\textbf{Dataset} & \textbf{Type} & \textbf{Area} (km$^2$) & \textbf{Length} (km) & \textbf{RGB Images} & \textbf{Resolution} & \textbf{Region} & \textbf{Year} \\
\midrule
SpaceNet \cite{van2018spacenet} & 2D & 5555 & 8676.6 & 24k & 650 × 650 &  Worldwide & 2018 \\
DeepGlobe \cite{demir2018deepglobe} & 2D & 2220 & - & 8k & 650 × 650 &  Worldwide & 2018 \\
Topo-boundary \cite{xu2021topo} & 2D & - & - & 25k & 1000 × 1000 &  USA & 2021 \\
nuScenes \cite{caesar2020nuscenes} & 3D & - & 242 & 1.4M & 1600 × 900 & Worldwide &  2020 \\
Argoverse2 \cite{wilson2argoverse} & 3D & - & 2220 & - & 2048 × 1550, 1550 × 2048 & USA & 2023 \\
OpenLane-V2 \cite{wang2024openlane} & 3D & - & - & 466k & 1600 × 900, 2048 × 1550, 1550 × 2048 & Worldwide & 2024 \\
\bottomrule
\end{tabular}
\label{tab:1}
\end{table*}

\begin{table*}[h]
\centering

\caption{Performance comparison of state-of-the-art aerial imagery-based methods on Topo-Boundary \cite{xu2021topo} dataset.}
\begin{threeparttable}  
\begin{tabular}{l|rrrrrrrrrrr}
\toprule
\multirow{2}{*}{\textbf{Method}} & \multicolumn{3}{c}{\textbf{Precision}} & \multicolumn{3}{c}{\textbf{Recall}} & \multicolumn{3}{c}{\textbf{F1 Score}} & \multirow{2}{*}{\textbf{APLS}} & \multirow{2}{*}{\textbf{TLTS}} \\
\cmidrule(lr){2-10}
 & 2.0 & 5.0 & 10.0 & 2.0 & 5.0 & 10.0 & 2.0 & 5.0 & 10.0 \\
\midrule
OrientationRefine \cite{batra2019improved} & 0.517 & 0.816 & 0.868 & 0.352 & 0.551 & 0.589 & 0.408 & 0.637 & 0.678 & 0.235 & 0.219 \\
Enhanced-iCurb \cite{xu2021topo} & 0.412 & 0.695 & 0.785 & 0.412 & 0.671 & 0.749 & 0.410 & 0.678 & 0.760 & 0.299 & 0.279 \\
Sat2Graph \cite{he2020sat2graph} & 0.460 & 0.484 & 0.604 & 0.128 & 0.240 & 0.293 & 0.159 & 0.304 & 0.374 & 0.037 & 0.030 \\
csBoundary \cite{xu2022csboundary} & 0.309 & 0.659 & 0.830 & 0.291 & 0.600 & 0.738 & 0.297 & 0.652 & 0.772 & 0.376 & 0.343 \\
\bottomrule
\end{tabular}
\label{tab:2}
\footnotesize{For all the metrics, larger values indicate better performance.}
\end{threeparttable}
\end{table*}

\begin{table}[h]
\centering
\caption{Performance comparison of state-of-the-art onboard sensors-based methods on OpenLane-V2 \cite{wang2024openlane} dataset.}
\begin{threeparttable}  
\begin{tabular}{l|rrrr}
\toprule
\multirow{2}{*}{\textbf{Method}} & \multicolumn{2}{c}{\textbf{Subset$\_$A}} & \multicolumn{2}{c}{\textbf{Subset$\_$B}} \\
\cmidrule(lr){2-3}\cmidrule(lr){4-5}
 & ${DET}_{l}$ & ${TOP}_{ll}$ & ${DET}_{l}$ & ${TOP}_{ll}$ \\
\midrule
% STSU \cite{can2021structured} & 12.7 & 0.5 & 8.2 & 0.0 \\
TopoNet \cite{li2023graph} & 28.5 & 4.1 & 24.3 & 2.5 \\
TopoMLP \cite{wu2023topomlp} & 28.3 & 7.2 & 26.6 & 7.6 \\
TSTGT \cite{rong2024driving} & 29.0 & 12.1 & 27.5 & 13.7 \\
SMERF \cite{luo2024augmenting} & 33.4 & 7.5 & - & - \\
LaneSegNet \cite{li2024lanesegnet} & 31.8 & 7.6 & - & - \\
TopoLogic \cite{fu2024topologic} & 34.4 & 23.4 & 25.9 & 15.1 \\
Topo2D \cite{li2024enhancing} & 29.1 & 22.3 & - & - \\
RoadPainter \cite{ma2024roadpainter} & 36.9 & 12.7 & 28.7 & 8.5 \\
\bottomrule
\end{tabular}
\label{tab:3}
\footnotesize{For all the metrics, larger values indicate better performance. Among all the methods, the feature extraction network adopted ResNet-50 \cite{he2016deep}, and the model is trained for 24 epochs.}
\end{threeparttable}
\end{table}

\subsection{Evaluation Metrics}

Three pixel-level metrics ($Precision$, $Recall$, and $F1$ score) \cite{xu2021topo} and two topology-level metrics ($APLS$ score and $TLTS$ score) \cite{xu2022csboundary} are commonly used road-level measures in aerial imagery-based methods. Additionally, lane-level metrics, such as perception metrics (${DET}_{l}$ score) and reasoning metrics (${TOP}_{ll}$ score) \cite{wang2024openlane} with a wide range of applications for topology between lane centerlines in onboard sensors-based methods. Consistent evaluation metrics for procedural modeling-based methods are lacking, so we do not discuss them here.

Pixel-level metrics are defined as follows:

\begin{equation}\label{eq:1}
Presicion = \frac{\left | \left \{ p|d(p,Q)<{\tau,\forall p\in P }\right \}\right |}{\left | P\right |}
\end{equation}

\begin{equation}\label{eq:2}
Recall = \frac{\left | \left \{ q|d(q,P)<{\tau,\forall q\in Q }\right \}\right |}{\left | Q\right |}
\end{equation}

\begin{equation}\label{eq:3}
F1~Score = \frac{2 \times Precision \times Recall}{Precision+Recall}
\end{equation}
where $P = {\left \{ {p}_{i}\right \}}_{i=1}^{{N}_{p}}$ and $Q = {\left \{ {q}_{j}\right \}}_{j=1}^{{N}_{q}}$ represent foreground pixels and ground-truth road-boundary graph in the rasterized prediction results, respectively.  $\left | \cdot \right |$ indicated the number of all elements within the set. $d\left ( e,S\right )$ denoted as the shortest Euclidean distance between element $e$ and a set $S$. Relax ratio is $\tau$, which reflects the level of error tolerance, generally setting to 2.0, 5.0, and 10.0.

The $APLS$ score is defined as follows:

\begin{equation}\label{eq:4}
APLS = 1-min\left ( 1,\frac{\left |l\left ( {g}_{1},{g}_{2}\right )-l\left ( {p}_{1},{p}_{2}\right ) \right |}{l\left ( {g}_{1},{g}_{2}\right )}\right )
\end{equation}
where ${g}_{1}$ and ${g}_{2}$ represent two vertices randomly selected in the ground-truth graph, calculate the shortest path between these two vertices, denoted as $l\left ( {g}_{1},{g}_{2}\right )$. In the same way, $l\left ( {p}_{1},{p}_{2}\right )$ represents the shortest path between two vertices ${p}_{1}$ and ${p}_{2}$ randomly selected in the predicted graph. The final $APLS$ score is the average of the $APLS$ of all selected vertex pairs.

The $TLTS$ score indicates the proportion of vertex pairs that are not \textit{too long or too short} among all vertex pairs. Specifically, an error tolerance threshold $\phi$ is defined (default value is 0.05), and the vertex pair is considered \textit{too long or too short} if they satisfy the following conditions:

\begin{equation}\label{eq:5}
\left |l\left ( {g}_{1},{g}_{2}\right )-l\left ( {p}_{1},{p}_{2}\right ) \right |>l\left ( {g}_{1},{g}_{2}\right )\cdot \phi 
\end{equation}

The ${DET}_{l}$ score is defined as follows:

\begin{equation}\label{eq:6}
{DET}_{l} = \frac{1}{\left | T\right |}\displaystyle\sum_{t\in T}{A}{P}_{t}
\end{equation}
where match thresholds $T = \left \{ 1.0,2.0,3.0\right \}$, which is based on the Fr{\'e}chet distance \cite{eiter1994computing}. The final ${DET}_{l}$ score is the mean of ${DET}_{l}$ of all three match thresholds.

Finally, the ${TOP}_{ll}$ score is:

\begin{equation}\label{eq:7}
	{TOP}_{ll} = \frac{1}{\left |V \right |}\displaystyle\sum_{v\in V}^{}\frac{\textstyle\sum_{{\hat{n}}^{\prime}\in {\hat{N}^{\prime}}\left ( v\right )}P\left ( {\hat{n}}^{\prime}\right )\mathrm{1}({\hat{n}}^{\prime}\in N\left ( v\right ))^{}}{\left | N\left ( v\right )\right |}
\end{equation}
where $N\left ( v\right )$ is a list in which every element is a neighbor of vertex $v$, ordered by confidence. $P\left ( v\right )$ is the precision of $i$-th vertex $v$ in the list $N\left ( v\right )$.

\subsection{Comparison Results}

Table \ref{tab:2} shows the quantitative results of the three types (segmentation-based, iterative graph construction, and graph growth) of aerial imagery-based methods on Topo-boundary \cite{xu2021topo} dataset, and the quantitative results for onboard sensors-based methods on OpenLane-V2 \cite{wang2024openlane} dataset are presented in Table \ref{tab:3}.

\section{challenges}
Despite the significant progress achieved in lane topology reasoning for HD mapping, several challenges remain to be addressed: (1) The construction of datasets dedicated to lane topology reasoning is hindered by high production costs, limiting the availability of large-scale, high-quality datasets. This scarcity restricts the diversity and generalizability of models trained for this task. More research efforts are required to develop accessible and robust datasets tailored to lane topology reasoning. (2) Although onboard sensors-based methods represent the state of the art, they typically involve models with substantial parameter sizes. This limits their direct deployment on autonomous vehicles or edge devices, which have stringent computational and energy constraints. Thus, it is crucial to develop lightweight models capable of real-time HD map construction. (3) Roadside and onboard camera data fusion for lane topology reasoning remain underexplored. Unlike onboard cameras, roadside sensors (e.g., road video surveillance) offer a complementary perspective for HD mapping, mitigating the problem of occlusion by surrounding vehicles and improving accuracy. However, limited literature exists on this line of research, signaling a need for further investigation.

% Despite the significant progress achieved in lane topology reasoning for HD mapping, several challenges remain to be addressed:
% \begin{itemize}
%     \item \textit{Dataset availability}. The construction of datasets dedicated to lane topology reasoning is hindered by high production costs, limiting the availability of large-scale, high-quality datasets. This scarcity restricts the diversity and generalizability of models trained for this task. More research efforts are required to develop accessible and robust datasets tailored to lane topology reasoning.
%     \item \textit{Model efficiency}. Although onboard camera-based methods represent the state of the art and exhibit outstanding performance, they typically involve models with substantial parameter sizes. This limits their direct deployment on autonomous vehicles or edge devices, which have stringent computational and energy constraints. The development of lightweight models capable of real-time HD map construction is crucial to enable practical applications in resource-constrained environments.
%     \item \textit{Exploration of roadside sensors}. Lane topology reasoning using images collected from roadside sensors remains an underexplored area. Unlike onboard cameras, roadside sensors (e.g., road video surveillance) offer a complementary perspective for HD mapping, mitigating the problem of occlusion by surrounding vehicles and improving accuracy. However, limited literature exists on this line of research, signaling a need for further investigation.
% \end{itemize}

\section{Conclusions}

Lane topology reasoning techniques have emerged as a critical component in HD mapping and autonomous driving applications, experiencing rapid evolution from early stages of procedural modeling to modern onboard sensors-based approaches. This survey has presented a comprehensive analysis of this progression, highlighting the significant advances achieved while identifying remaining challenges and opportunities for future research. Contemporary techniques have demonstrated superior performance in capturing complex lane topologies. The integration of various data sources - including camera imagery, LiDAR point clouds, and existing SD maps - has further enhanced the robustness and accuracy of these systems. However, several critical challenges persist. The limited availability of large-scale, high-quality datasets continues to constrain model development and validation. Additionally, the computational demands of current architectures pose challenges for real-time applications in resource-constrained environments. For future research in the field, a promising direction is to create lane topology by adopting roadside infrastructure-assisted vehicle perception techniques.

\section{Acknowledgements}
This work was sponsored by Beijing Nova Program (No. 20240484616). 

\balance
\bibliographystyle{IEEEtran}
\bibliography{IEEEexample}
\end{document}